%% file: main.tex
\documentclass[10pt,twocolumn,letterpaper]{article}

\usepackage{cvpr}
\usepackage{times}
\usepackage{epsfig}
\usepackage{graphicx}
\usepackage{amsmath}
\usepackage{amssymb}
\usepackage[dvipsnames]{xcolor}
\usepackage{subfigure}
\usepackage[titletoc,title]{appendix}

\DeclareMathOperator*{\argmin}{arg\,min}

\newif\ifdraft\drafttrue
\ifdraft

\newcommand\sd[1]{{\footnotesize \color{green}[#1 - \textbf{Sangdoo}]}}
\newcommand\hj[1]{{\footnotesize \color{ForestGreen}[#1 - \textbf{Hyung Jin}]}}
\newcommand\jy[1]{{\footnotesize \color{magenta}[#1 - \textbf{Jin Young}]}}
\newcommand\yd[1]{{\footnotesize \color{cyan}[#1 - \textbf{Yiannis}]}}
\else

\newcommand\jw[1]{}
\newcommand\sd[1]{}
\newcommand\hj[1]{}
\newcommand\jy[1]{}
\newcommand\yd[1]{}
\fi

\usepackage[pagebackref=true,breaklinks=true,colorlinks,bookmarks=false]{hyperref}

\cvprfinalcopy 

\def\cvprPaperID{1355} 

\ifcvprfinal\fi
\begin{document}

\title{Variational Autoencoded Regression: \\High Dimensional Regression of Visual Data on Complex Manifold}

\author{YoungJoon Yoo\textsuperscript{1} \quad Sangdoo Yun\textsuperscript{2} \quad Hyung Jin Chang\textsuperscript{3} \quad Yiannis Demiris\textsuperscript{3} \quad Jin Young Choi\textsuperscript{2}\\
\textsuperscript{1}Graduate School of Convergence Science and Technology, Seoul National University, South Korea\\
\textsuperscript{2}ASRI, Dept. of Electrical and Computer Eng., Seoul National University, South Korea\\
\textsuperscript{3}Personal Robotics Laboratory, Department of Electrical and Electronic Engineering\\ Imperial College London, United Kingdom\\
{\tt\small 
\textsuperscript{1}yjyoo3312@gmail.com 
\textsuperscript{2}\{yunsd101, jychoi\}@snu.ac.kr 
\textsuperscript{3}\{hj.chang, y.demiris\}@imperial.ac.uk}
}


\maketitle

\input{0_Abstract.tex}
\input{1_Introduction.tex}
\input{2_RelatedWork.tex}
\input{3_ProposedMethod.tex}
\input{4_Experiment.tex}
\input{5_Conclusion.tex}

\section{Acknowledgment}
This work was partly supported by the ICT R\&D program of MSIP/IITP (No.B0101-15-0552, Development of Predictive Visual Intelligence Technology), the SNU-Samsung Smart Campus Research Center at Seoul National University, 
EU FP7 project WYSIWYD under Grant 612139 and the BK 21 Plus Project.
We thank the NVIDIA Corporation for their GPU donation.

{\small
\bibliographystyle{ieee}
\bibliography{egbib}
}

\input{6_Appendix.tex}

\end{document}

%% file: 0_Abstract.tex
\begin{abstract}
This paper proposes a new high dimensional regression method by merging Gaussian process regression into a variational autoencoder framework. In contrast to other regression methods, the proposed method focuses on the case where output responses are on a complex high dimensional manifold, such as images. Our contributions are summarized as follows: (i) A new regression method estimating high  dimensional image responses, which is not handled by existing regression algorithms, is proposed. (ii) The proposed regression method introduces a strategy to learn the latent space as well as the encoder and decoder so that the result of the regressed response in the latent space coincide with the corresponding response in the data space. (iii) The proposed regression is embedded into a generative model, and the whole procedure is developed by the variational autoencoder framework. We demonstrate the robustness and effectiveness of  our method through a number of experiments on various visual data regression problems.
\end{abstract}

%% file: 1_Introduction.tex
\section{Introduction}

\begin{figure}
\begin{center}
   \includegraphics[width=0.95\linewidth]{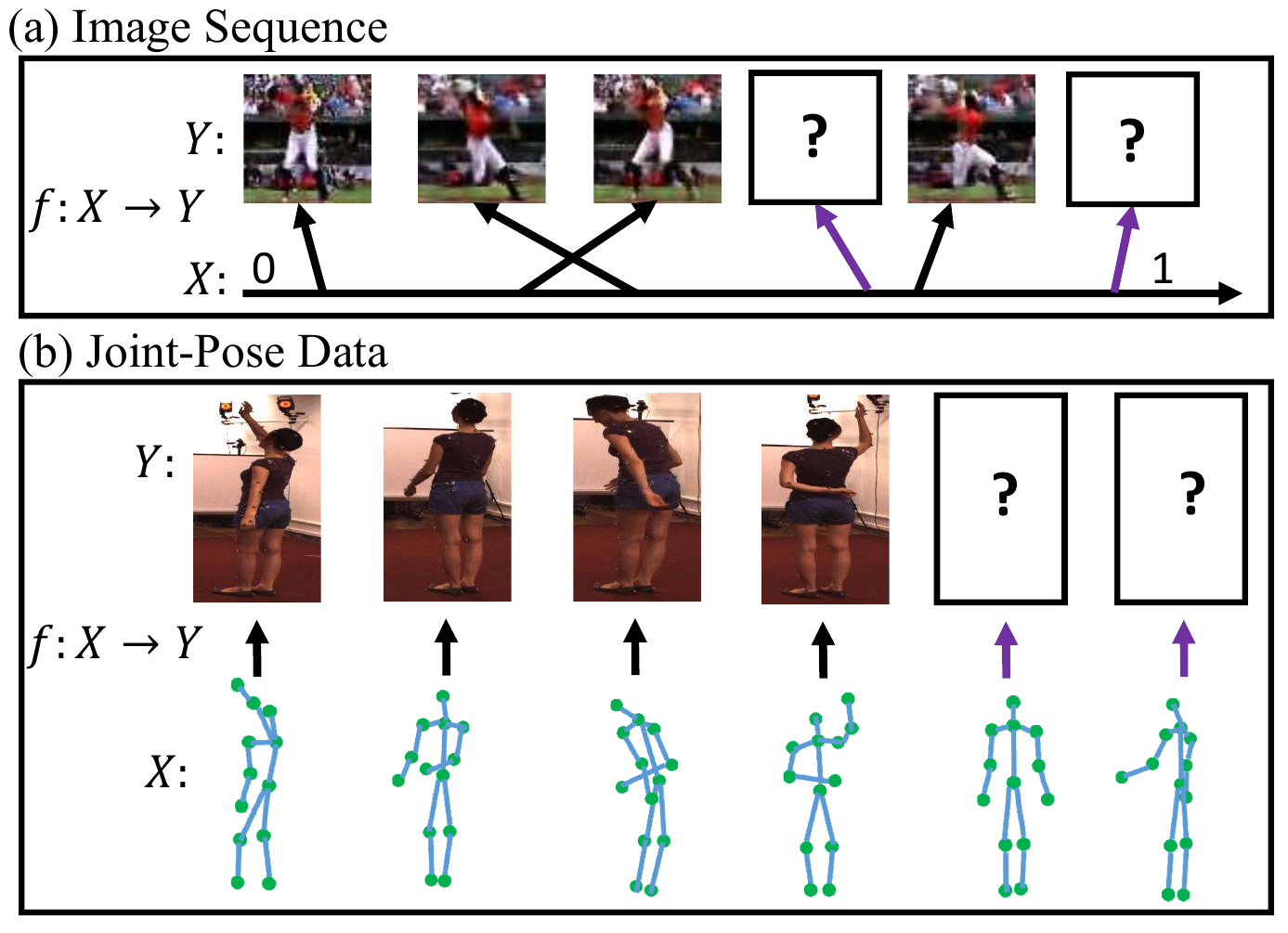}
\end{center}
\vspace{-3mm}
   \caption{Examples of paired data in vision applications.
   (a) For the image sequence, the domain can be defined by the space representing relative orders of image sequences. 
   (b) For joint-pose data pairs, the joint vector space can be a possible domain.}
   
\label{fig:intro}
\end{figure}

Regression of paired input and output data with an unknown relationship is one of the most crucial challenges in data analysis. 
In diverse research fields, such as trajectory analysis, robotics, the stock market,  etc.~\cite{he2011single,yang2002support,yang2002support,lo1988stock,kimoto1990stock}, target phenomena are interpreted as a form of paired input / output data.
In these applications, a regression algorithm is usually used to estimate the unknown response for a given input by using the information obtained from the observed data pairs. 
Many vision applications can also be expressed as such input / output data pairs. 
For example, in Fig.~\ref{fig:intro}~(a), the sequence of the motion images can be described by input / output paired data, where the input can be defined as the relative order and the output response is defined as the corresponding image.
The motion capture data and their corresponding images in Fig.~\ref{fig:intro}~(b) are another example.
The input data are 3D joint positions, and their responses will be the corresponding posture images.
If we can model the implicit function representing the given image data pairs via regression, we can estimate unobserved images that correspond to the input data.

However, applying existing multiple output regression algorithms~\cite{alvarez2009sparse,alvarez2010efficient,alvarez2011computationally,swersky2013multi} to these kinds of visual data applications is not straightforward, because the visual data are usually represented in high dimensional spaces.
In general, high dimensional visual data (such as image sequences) are difficult to be analyzed with classical probabilistic methods because of their limited modeling capacity~\cite{hinton2002training, hinton2006fast}.
Thus, regression of visual data to estimate visual responses requires a novel approach.

In handling high dimensional complex data, recent attempts~\cite{goodfellow2014generative,li2015generative,salakhutdinov2009learning,kingma2013auto,rezende2014stochastic} using a deep generative network, such as the variational autoencoder (VAE)~\cite{kingma2013auto}, have achieved significant success in reconstructing images.
In VAE, a latent variable is defined to embed the compressed information of the data, and an encoder is trained to map a data space into its corresponding latent space. 
A decoder is also trained to reconstruct images from a sampled point in the latent space. 
The projection for the input data into the latent space (via the encoder) captures the essential characteristics of the data and allows execution of the regression task using a much lower dimensional space. 
The regression in the latent space is done together with the training of the encoder and decoder.
However, a naive combination of regression and the VAE is not particularly effective 
because the decoder and the latent space are not designed in a  way that permits the result of the regressed response in latent space, and the corresponding response in data space, to coincide.
Therefore, a new method to simultaneously train the latent space and the encoder/decoder is required to achieve  coincidence between the regressed latent vector and the reconstructed image.

In this paper, we solve this problem by combining the VAE~\cite{kingma2013auto} and Gaussian process regression~\cite{rasmussen2006gaussian}.
The key idea of this work is to do regression in the latent space instead of the high-dimensional image space.
The proposed algorithm generates the latent space that compresses the information of both the domain and output image using the VAE, and projects the data pairs to the latent space. 
Then, regression is conducted for the projected data pairs in latent space, and the decoder is trained so that the regression in latent space and image space coincide. 
To the best of our knowledge, it is the first attempt to apply the VAE framework to solve the regression problem. 
The whole process, including the loss function, is designed as the generative model, and a new mini-batch composition method is presented for training the decoder to satisfy our purpose.

All connection parameters of the encoder / decoder are inferred by the end-to-end stochastic gradient descent method as described in~\cite{kingma2014adam}.
The proposed regression method is validated with two different examples: sports sequences and motion image sequences with skeletons. 
The first example presents a regression case of simple domain to complex codomain, and the second example presents the complex domain to complex codomain case.


%% file: 2_RelatedWork.tex
\section{Related Work}
\noindent
\textbf{Deep Generative Network: }
Classical probabilistic generative models~\cite{rasmussen1999infinite,blei2003latent,wei1998exponential,holland1981exponential,maceachern1998estimating,teh2012hierarchical,baum1966statistical} have proven to be successful in understanding diverse unsupervised data, but their descriptive ability is insufficient to fully explain complex data such as images~\cite{hinton2006fast}.
Recently, as in other works in the vision area~\cite{he2016deep,nam2016learning}, deep layered architectures have been successfully applied to solve this problem with powerful data generation performance. 
Among these architectures, generative adversarial network (GAN)~\cite{goodfellow2014generative} and generative moment matching networks (GMMN)~\cite{li2015generative} directly learn the generator that maps latent space to data space.
Meanwhile, the variants of the restricted Boltzman machine (RBM) ~\cite{hinton2002training,hinton2006fast,salakhutdinov2009deep,salakhutdinov2009learning} and probabilistic autoencoders~\cite{kingma2013auto,makhzani2015adversarial,germain2015made} learn the encoder that defines the map from data to latent space and the generator (decoder) simultaneously. 
The former methods, and especially variants of GAN~\cite{goodfellow2014generative,denton2015deep,radford2015unsupervised}, are reported to describe the edges of generated images more sharply than the latter methods.
However, the applicability of these methods is restricted due to the difficulty of discovering the relationships between data and latent space.
This innate nature makes it difficult to use adversarial networks for designing the regression.
Therefore, this paper adopts the variational autoencoder framework~\cite{kingma2013auto}, which is also more suitable than RBM families to expand the regression model. 
\begin{figure*}
\begin{center}
   \includegraphics[width=0.9\linewidth]{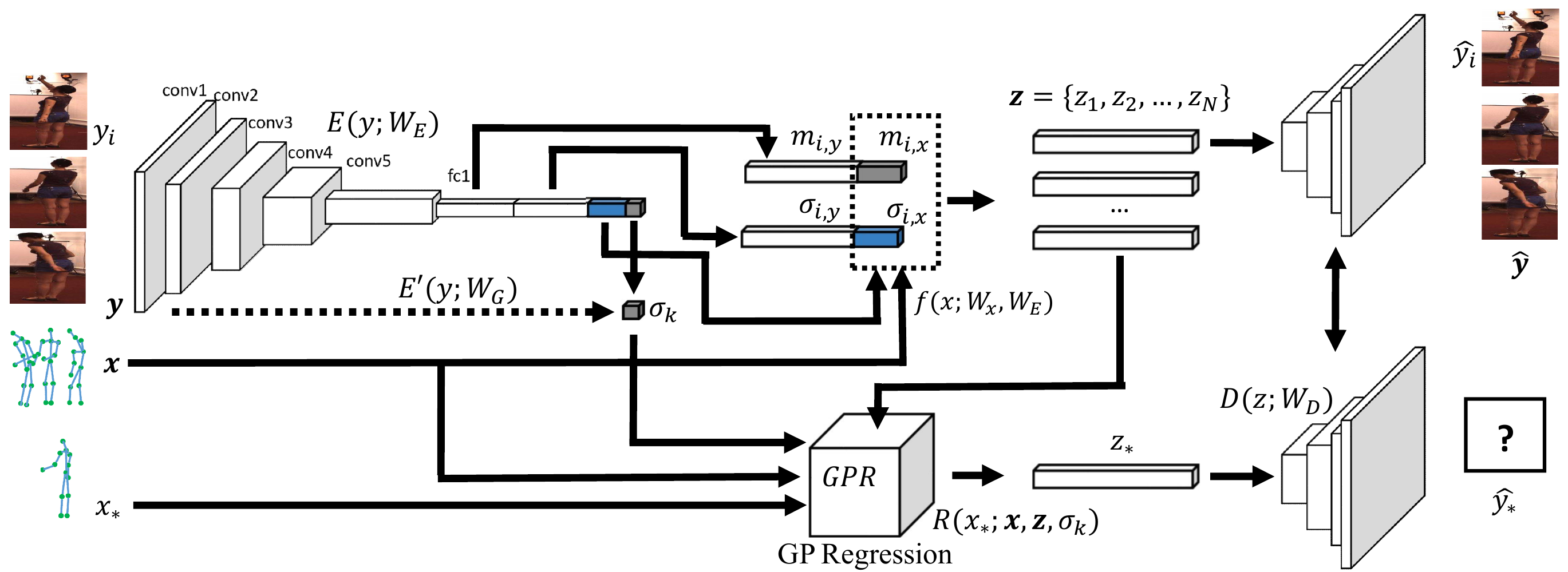}
   \vspace{-3mm}
\end{center}
   \caption{Overall scheme of the proposed method. For observed data pairs $\textbf{x} = \{x_1,x_2,\cdots,x_N\}$ and $\textbf{y} = \{y_1,y_2,\cdots,y_N\}$, the proposed autoencoder reconstructs $\hat{y_i}\simeq y_i$ as shown in the top right. 
   For the unobserved $y_*$ to given $x_*$, 
   it is impossible to obtain $z_*$ by using an encoder with $W_E$, because we do not have information about $y_*$. Thus to estimate $y_*$, we obtain $z_*$ using regression from $\textbf{x}, \textbf{z}$ and $x_*$, and estimate the response $\hat{y}_*$ from $z_*$. 
   }
\label{fig:overview}
\vspace{-3mm}
\end{figure*}

\noindent
\textbf{Variational Autoencoder: }
Since Kingma \etal~\cite{kingma2013auto} first published the variational autoencoder (VAE), numerous applications~\cite{pu2016variational,yan2015attribute2image} have been presented to solve various problems. 
Yan \etal ~\cite{yan2015attribute2image} proposed conditional VAE in order to generate the image conditioned on the attribute given in the form of sentences.
Furthermore, recent work~\cite{gregor2015draw,krishnan2015deep,goroshin2015learning} has demonstrated  that a sequence in latent space would be mapped back to the sequence of data. 
Hence these methods embedded dynamic models such as recurrent neural networks~\cite{gregor2015draw} and the Kalman filter~\cite{krishnan2015deep,kalman1960new} into the VAE framework. 
These algorithms~\cite{gregor2015draw,krishnan2015deep,krishnan2015deep,goroshin2015learning} successfully show the ability of dynamic models in a latent space to capture the temporal changes of relatively simple objects in images.
In this paper, we apply the VAE for the regression task in a relatively complex manifold.

\noindent
\textbf{Regression: }
Regression of paired data is theoretically well established and analytic solutions for the infinite dimension of the basis function~\cite{anzai2012pattern,rasmussen2006gaussian} have been derived for the last century. 
In non-parametric cases, Gaussian process~\cite{rasmussen2006gaussian,lawrence2004gaussian} provides a general solution by expanding Bayesian linear regression  using kernel metrics and a Gaussian process prior. 
By using this method, we can estimate the output data as a Gaussian posterior composed of given data pairs and input data to the unobserved target outputs. 
However, applying the algorithms for the high-dimensional output data is difficult because the kernel metric has limited capacity to express the complicated high dimensional data. 
The variants of multiple output regression algorithms~\cite{alvarez2011computationally,alvarez2010efficient,alvarez2009sparse,swersky2013multi} are proposed to deal with multi-dimensional output responses.
Still, these algorithms focus on handling relatively low dimensional output responses and are not able to sufficiently describe complicated data, such as that of an image.
In this paper, we construct a regressed latent space by using a variational autoencoder to handle complex data.


%% file: 3_ProposedMethod.tex
\section{Proposed Method}
\subsection{Overall Scheme}
\label{sec:overall_scheme}
Given the target data pair $(x_i,y_i),i=1,\ldots,N, x_i\in\mathcal{X}, y_i\in\mathcal{Y}$, our goal is to find the unknown response $y_*$ for a new input $x_*$.
In this paper, the response $y\in\mathcal{Y}$ is defined as an image and the corresponding input $x\in\mathcal{X}$ is defined accordingly based on the applications as in Fig.~\ref{fig:intro}.


As shown in Fig.~\ref{fig:overview}, 
for the observed data pairs $(x_i,y_i),i=1,\cdots,N$, the encoder/decoder produces $\hat{y_i}$ which is the reconstruction of an observed image $y_i$. 
For the observed data, the encoding network $E(\cdot)$ produces mean and variance for a part of the latent vector $z_i$, that is, $[m_{i,y},\sigma_{i,y}] = E(y_i;W_E)$ which compresses $y_i$ to a latent variable with Gaussian mean $m_{i,y}$ and variance $\sigma_{i,y}$.
The remaining part of $z_i$ is modeled by $[m_{i,x}, \sigma_{i,x}]=f(x_i,W_x)$ which represents mean and variance of $z_i$. 
Thus, the Gaussian distribution of $z_i$ is described by $m_i = [m_{i,y}, m_{i,x}]$ and $\sigma_i = diag[\sigma_{i,y}, \sigma_{i,x}]$.
For the unobserved image $y_*$ for a newly given $x_*$, the proposed method produces $\hat{y}_*$, which is an estimator of $y_*$.


Using $z_i$ sampled from $\mathcal{N}(m_i,\sigma_i)$, the decoding network reconstructs the output response $\hat{y}_i$, that is, $\hat{y_i} = D(z_i;W_D)$. 
Note that if $(W_E, W_x, W_D)$ is well trained by the training scheme in Section~\ref{sec:inference}, $\hat{y_i}$ should be similar to $y_i$.
However, for an unobserved $y_*$ to a given $x_*$, it is impossible to obtain $z_*$ from $E(\cdot;W_E)$ because we do not have any information on $y_*$. 
To estimate $y_*$, we obtain $z_*$ by using regression from $\textbf{x}=\{x_1,x_2,\cdots,x_N\}, \textbf{z}=\{z_1,z_2,\cdots,z_N\}$ and $x_*$.
For this regression, $z_i$ is sampled from $\mathcal{N}(m_i,\sigma_i)$ for each observed response $y_i\in\textbf{y}=\{y_1,y_2,\cdots,y_N\}$.
Then, we estimate $z_*$ using Gaussian process (GP) regression $z_* \sim R(x_*;\textbf{x},\textbf{z},\sigma_k)$ to be described in Section~\ref{sec:VAE_R}, where $\sigma_k$ is a kernel parameter of the GP regression, which can be produced by an additional encoder $\sigma_k = E'(\textbf{y}, W_G)$; with $\sigma_k =[\sigma_{k,1}, \cdots, \sigma_{k,N}]$. 
In this paper, for computational simplicity, we combine this kernel encoder with the encoder network $E(y;W_E)$ and change the outputs into $[m_{i,y}, \sigma_{i,y}, \sigma_{i,k}] = E(y_i, W_E)$.

After $z_*$ is estimated, the response $\hat{y}_*$ is reconstructed from $z_*$ by using the decoding network $D(z_*;W_D)$.
Note that the $D(z_*;W_D)$ should reconstruct not only $\hat{y}$ from the $z_i$ sampled by $\mathcal{N}(m_i,\sigma_i)$, but also $y_*$ from the $z_*$ which is the regression result obtained from $x_*,\textbf{x},$ and $\textbf{y}$.
The whole procedure is designed as a generative framework with joint distribution $p(x_*,y_*,\textbf{x},\textbf{y},W_E,W_x,W_D)$, and hence can be derived by the VAE algorithm.
\subsection{Variational Autoencoded Regression}
\label{sec:VAE_R}

\begin{figure}
\begin{center}
   \includegraphics[width=0.9\linewidth]{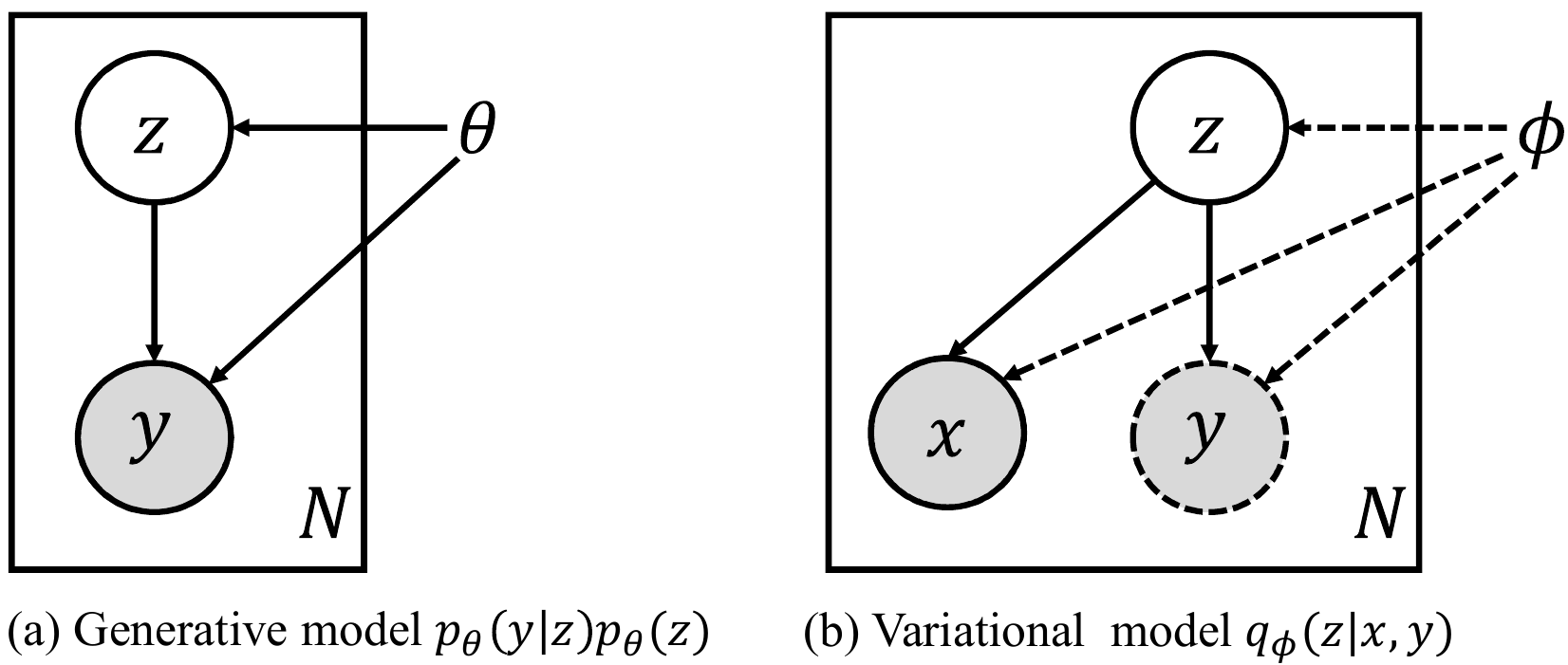}
    \vspace{-2mm}
\end{center}
\vspace{-3mm}
   \caption{The directed graphical model of the proposed method.
   (a) Generative model for $y$ and the latent variable $z$.
   (b) Variational distribution to approximate the posterior $p_\theta(z|y)$ of the generative model.
   $y$ is not observed for the newly given input $x=x_*$.}
\label{fig:directed_graph}
 \vspace{-3mm}
\end{figure}

The proposed scheme (depicted in Fig.~\ref{fig:overview}) is derived from the directed graph model in Fig.~\ref{fig:directed_graph}.
The diagram in Fig.~\ref{fig:directed_graph}~(a) represents the generative model describing a typical reconstruction problem, and the diagram in Fig.~\ref{fig:directed_graph}~(b) is the variational model which not only approximates the generative model in Fig.~\ref{fig:directed_graph}~(a), but also performs the regression for the estimation of unobserved $y$ by utilizing an information variable $x$ related to $y$.
The joint distribution $p_\theta(y,z)$ can be expressed by the likelihood function $p_\theta(y|z)$ and the prior distribution $p_\theta(z)$, where $\theta$ refers to the  set of all parameters related to the generation of the response $y\in\mathcal{Y}$ from the latent variable $z$.
In our method, the prior distribution of $z$ is defined as zero mean Gaussian distribution, as in typical variants of VAE~\cite{yan2015attribute2image,kingma2013auto}.
Also, the likelihood function $p_\theta(y|z)$ depicts the decoding process in the proposed scheme.
Below, it is shown that $\theta$ is realized by the parameter $W_D$ of the decoding network.

Once the joint distribution $p_\theta(y,z)$ is defined, the posterior $p_\theta(z|y)$ can be theoretically derived from the Bayes theorem, but the calculation is intractable. 
Therefore, the variational distribution $q_\phi(z|x,y)$ is introduced to approximate the true posterior distribution $p_\theta(z|y)$. 
Unlike $p_\theta(z|y)$, $x$ is introduced for the variational distribution $q_\phi(z|x,y)$ to sample $z_* \sim R(x_*;\textbf{x},\textbf{z},\sigma_k)$, which is the result of the GP regression for the unknown $y_*$.
$q_\phi(z|x,y)$ represents the overall encoding procedure generating the latent variable $z$ from the input data pair $(x,y)$ and correspondingly, the variational
parameter $\phi$ is realized by the parameters $W_E, W_x$ as described in Section~\ref{sec:overall_scheme}.
Importantly, $q_\phi(z|x,y)$ should be able to explain both cases: 1) an observed image $y_i$, and 2) an unobserved image $y_*$ which requires regression as mentioned previously. 
For the first case, the variational distribution is defined as $z_i\sim q_\phi(z|x=x_i,y=y_i)$.
For the latter case, the variational distribution is defined as $z_{*}\sim q_\phi(z|x = x_{*},y\in\varnothing)$ which represents the GP regression procedure for estimating latent $z_*$ for the input $x_*$.

In order to estimate the parameters $\theta$ and $\phi$ which minimize the distance between $p_\theta(z|y)$ and $q_\phi(z|x,y)$, we minimize the Kullback$-$Leibler divergence $D_{KL}(p_\theta(z|y)||q_\phi(z|x,y))$.
Following the derivation in \cite{beal2003variational,kingma2013auto}, the minimization procedure $\{\theta^*,\phi^*\} = \argmin_{\{\theta,\phi\}}D_{KL}(p_\theta(z|y)||q_\phi(z|x,y))$ is converted to $\{\theta^*,\phi^*\} = \argmin_{\{\theta,\phi\}}L(\theta,\phi)$, where

\vspace{-3mm}
\begin{eqnarray}
\begin{aligned}
\label{eq:RVAE}
L(\theta,\phi)=&-D_{KL}(q_\phi(z|x,y)||p_\theta(z))\\
+&\sum^{N}_{i=1}{\log{p_\theta(y_i|z_i)}}+\sum^{M}_{j=1}{\log{p_\theta(\hat{y}_{*j}|z_{*j})}}.\\
\end{aligned}
\vspace{-3mm}
\end{eqnarray}
$z_{*,j}$ and $y_{*,j}$ represents the $M$ number of latent codes and output responses for $x_{*,j}, j=1,\cdots,M$, to be regressed.
In (\ref{eq:RVAE}), $\hat{y}_{*,j} = D(z_{*,j};W_D)$ is the reconstructed response from $z_{*,j}$ by the decoding network as depicted in Fig.~\ref{fig:overview}.
The parameters $\theta$ and $\phi$ are realized by the connection parameters of the encoding network with regression for $q_{\phi}(z|x,y)$, and the decoding network for $p_\theta(z|y)$ (see Section~\ref{sec:VAE_R_description}).
To minimize the loss in (\ref{eq:RVAE}), we propose a method for mini-batch learning (see Section~\ref{sec:inference}). The Adam optimizer~\cite{kingma2014adam} is used for stochastic gradient descent training.

\subsection{Model Description}
\label{sec:VAE_R_description} 
For the encoding part, we define $q_\phi(z|x,y)$ which maps the data pair $(x, y)$ into the latent space $\mathcal{Z}$.
For both, observed and unobserved images, $q_\phi(z|x,y)$ is defined by Gaussian distribution as in (\ref{eq:q_define}) and it enables us to analytically solve the 
KL-divergence term  $D_{KL}(q_\phi(z|x,y)||p_\theta(z))$ in (\ref{eq:RVAE}) following~\cite{kingma2013auto}:
\begin{equation} 
q_\phi(z|x,y) = \mathcal{N}(z|m(x,y), \sigma(x,y)).
\label{eq:q_define}
\end{equation}
The variational parameter $\phi$ consists of the Gaussian mean function $m(x,y)$ and the variance function $\sigma(x,y)$.
The $m(x,y)$ and $\sigma(x,y)$ are produced in different ways depending on the input data.
When the input data is given by $x=x_i \in \textbf{x}$, the encoder yields $m(x,y)=[m_{i,y},m_{i,x}]$ and $\sigma(x,y)=diag[\sigma_{i,y},\sigma_{i,x}]$, 
where $diag[\cdot]$ refers to a diagonal matrix.
When the input data is given by $x=x_{*,j}\in\textbf{x}_*$, $m(x,y)$ and $\sigma(x,y)$ are determined by the mean and variance $(m_{*,j},\sigma_{*,j})$ estimated by GP regression from $\textbf{z}$, $\textbf{x}$ and $x_{*,j}$, where 
\begin{equation}
\label{eq:Rmeanvar}
m_{*,j} = K_{*,j}K^{-1}\textbf{Z},~\sigma_G = (K_{**,j} - K_{*,j}K^{-1}K^T_{*,j})I.
\end{equation}
$\textbf{Z}$ refers to the matrix $[z_1;z_2;\cdots;z_N]\in\mathcal{R}^{N\times D}$, and $I\in\mathcal{R}^{D\times D}$ is the identity matrix, where $D$ is the dimension of  $z\in\mathcal{Z}$.
The matrices $K, K_{**,j}$ and $K_{*,j}$ are defined as
\begin{align}
&K = \begin{bmatrix}
k(x_1,x_1)  & \cdots & k(x_1,x_N)\\
\vdots      & \ddots & \vdots     \\
k(x_N,x_1)  & \cdots & k(x_N,x_N)\\
\end{bmatrix},\\
&K_{**,j} = k(x_{*,j},x_{*,j}),\\
&K_{*,j} = {[k(x_{*,j},x_1), k(x_{*,j},x_2),  \cdots , k(x_{*,j}, x_N)]}.
\label{eq:kernel_vector}
\end{align}
For the kernel $k(\cdot,\cdot)$, we use a simplified version of SE-kernel~\cite{rasmussen2006gaussian}, where $k(x_i, x_j)= \sqrt{\sigma_{i}\sigma_{\textbf{}j}}\exp{{||x_i-x_j||}^2}$.
Eventually, the variational parameter $\phi$  is realized by the weight matrices $(W_x, W_E)$ of the encoder network.
In summary, for the given data $\textbf{x}, \textbf{y},$ and $\textbf{x}_*$, $q_\phi(z|x,y)$ in (\ref{eq:q_define}) is given as 
\begin{equation}
q_\phi(z|x,y) = \begin{cases}
\mathcal{N}(m_i, \sigma_i) &x=x_i, y=y_i.\\
\mathcal{N}(m_{*,j}, \sigma_{*,j}) &x=x_{*,j}, y\in\varnothing.
\end{cases}
\label{eq:meanvar}
\end{equation}
\begin{figure}
\begin{center}
   \includegraphics[width=0.96\linewidth]{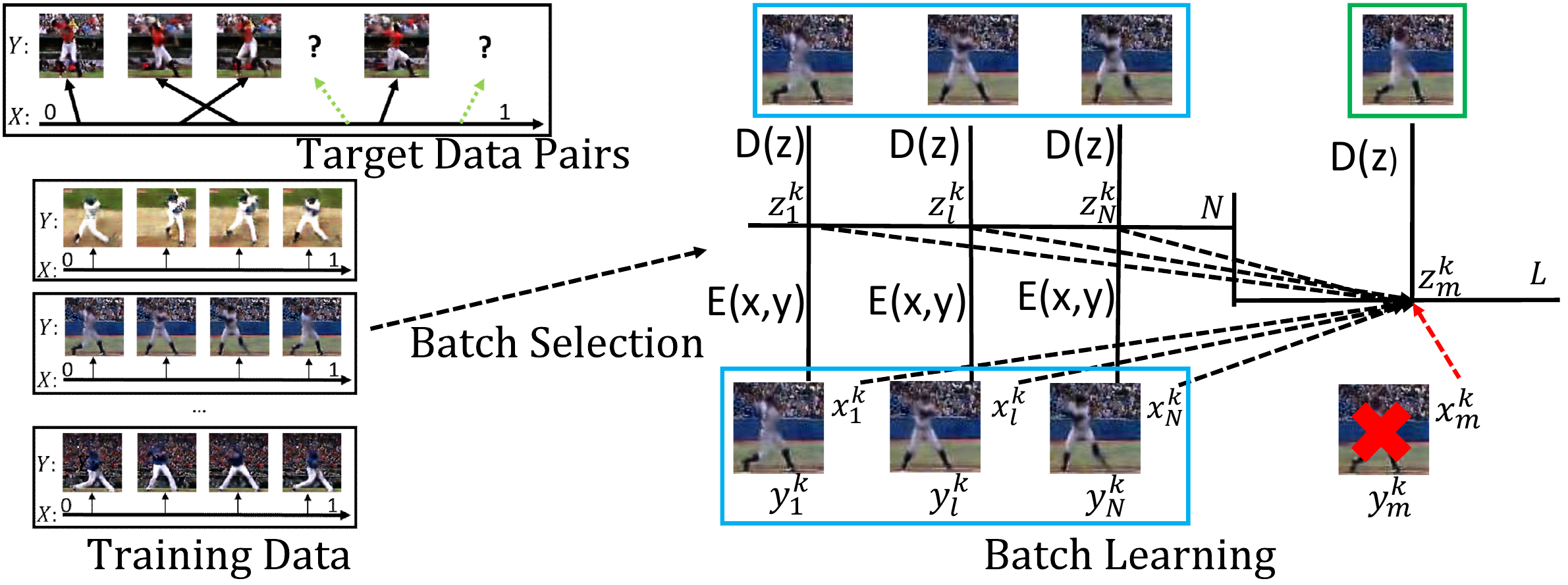}
\end{center}
\vspace{-3mm}
   \caption{Training strategy of the proposed method. The mini-batch is generated from the sampled training data sequences. }
\label{fig:BatchLearning}
\end{figure}

For the decoding procedure, we define the likelihood function $p_\theta(y|z) = p(y|D(z;W_D))$, where 
$p(y|D(z;W_D))$ is defined as a Gaussian distribution with mean $D(z;W_D)$ and fixed variance.
Since the prior of $z$ is defined with zero mean Gaussian and identity covariance matrix, 
the weight $W_D$ represents the generative model parameter $\theta$.
Correspondingly, the meanings of the second term and third term in (\ref{eq:RVAE}) are interpreted as follows. 
Since the negative log-likelihood $(-\log(p_\theta(y|z)))$ is defined as $l_2$ distance ${||y-D(z;W_D)||}^2$ in our algorithm, the second term represents the reconstruction error for the given data pair $(x_i,y_i)$ to $y_i$, and the third term denotes the estimation error for $y_{*,j}$ via regression from the given input data $x_{*,j}$ and the observed data $(x_i,y_i),i=1,\ldots,N$.
\subsection{Training } 
\label{sec:inference}
To train the parameters of the proposed model, a sufficient number of the training datasets is required.
In our algorithm, a total of $V$ different training sequences $(x^v_i, y^v_i), v=1,\ldots,V, i = 1\ldots,N_v$ are used, as shown in Fig.~\ref{fig:BatchLearning}. 
These training data pairs share similar semantics to the target~(test) data pair $(x_i,y_i)$. 
If the target data pair is a golf swing sequence, the training data pairs will be different golf swing sequences obtained in different situations.
Once the parameters are trained by the training dataset consisting of diverse golf swings, the proposed method can complete the target image sequence via regression from the incomplete test sequence on a golf swing.
After training the model with the mini-batch, we fine-tune the parameters with observed data pairs in target regression.

\noindent
\textbf{Mini-Batch Training: }
The work in \cite{bengio2009curriculum} reports that the composition of the mini-batch is critical when using variants of stochastic gradient descent methods~\cite{kingma2014adam,duchi2011adaptive} to train the parameters. 
To generate the batch, in this paper, $K$ sequences of a total $V$ sequences are randomly selected. 
For each selected training sequence $k=1\cdots K$, we randomly pick $L$ data pairs $(x^k_l,y^k_l), l=1,...,L$, where $L = (M+N)$.
For the earlier $N$ data pairs $(x^k_n,y^k_n), n=1,\cdots,N$, we get the latent space vector $z^k_n$ from the encoder function $E(y^k_n;W_E)$, and $f(x^k_n;W_x)$ to train $W_E, W_x$, and $W_D$.
Alternatively, for the latter $M$ data pairs $(x^k_m,y^k_m), m=(N+1),\cdots,L,$ we obtain the latent $z^k_m$ by regression (Section~\ref{sec:VAE_R_description}) from $\{z^k_1,\cdots,z^k_N\}$, $\{x^k_1,\cdots,x^k_N\}$ and $x^k_m$.
The responses $\{y^k_{N+1},\cdots,y^k_L\}$ are assumed to be unknown in the encoding process.
This data set is used to train the decoder network $D(z;W_D)$ to reconstruct the proper responses not only for the $z^k_n$ from the data pair $(x^k_n,y^k_n)$, but also for the $z^k_m$ which are obtained from the regression.
The corresponding loss from the estimated $\hat{y}^k_m$ and the actual $y^k_m$ refers to the the third term in (\ref{eq:RVAE}).
We note that it is possible to calculate the loss term because $y_m^k$ can be used as ground truth regression response.
After constructing the batch, the stochastic gradient~\cite{kingma2014adam} for the batch is calculated to train all parameters.
\begin{figure}
\begin{center}
   \includegraphics[width=0.96\linewidth]{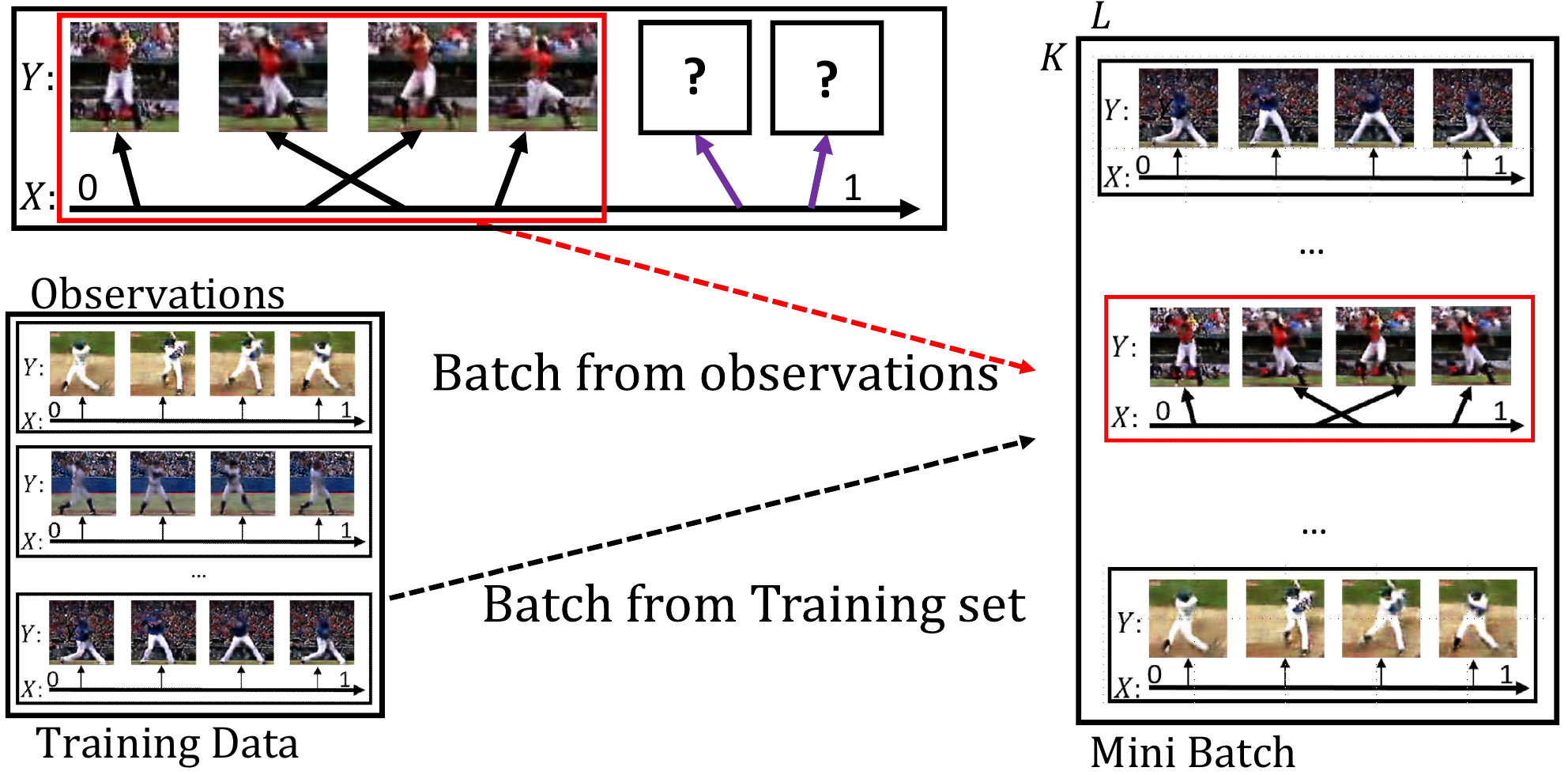}
\end{center}
\vspace{-3mm}
   \caption{Batch generation for fine-tuning. The batch is composed of observed data pairs (red) and sampled data pairs in training dataset.}
\label{fig:post_proc}
 \vspace{-3mm}
\end{figure}

\begin{figure*}
\begin{center}
   \includegraphics[width=0.99\linewidth]{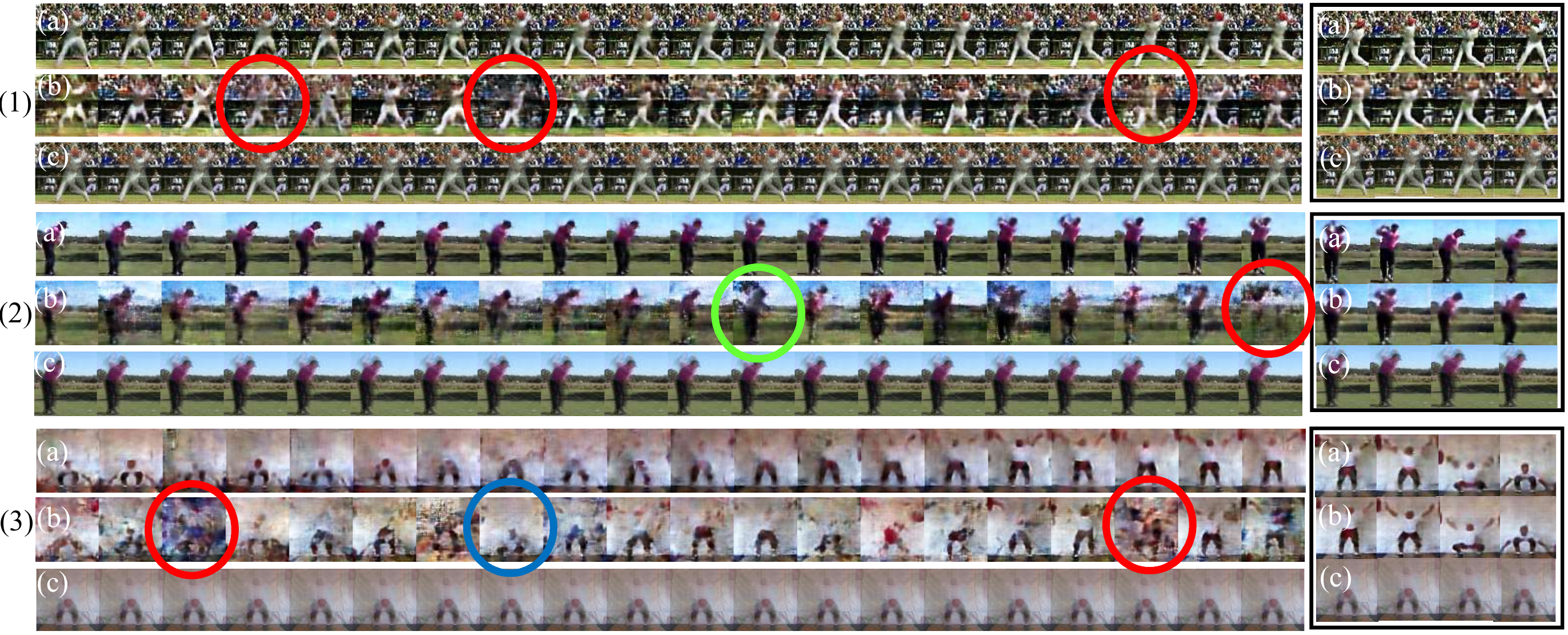}
\end{center}
   \vspace{-4mm}
   \caption{Qualitative Results on regression from the sport dataset (best viewed in color). The row (a) in each sport represents the proposed regression results. The images in rows (b) result from the regression with R-VAE. Row (c) is the result from MOGP~\cite{alvarez2011computationally}. The results on the right indicate the samples of reconstruction results for observed images.}
\label{fig:exp_sports_1}
\vspace{-3mm}
\end{figure*}
\begin{figure}
\begin{center}
   \includegraphics[width=0.99\linewidth]{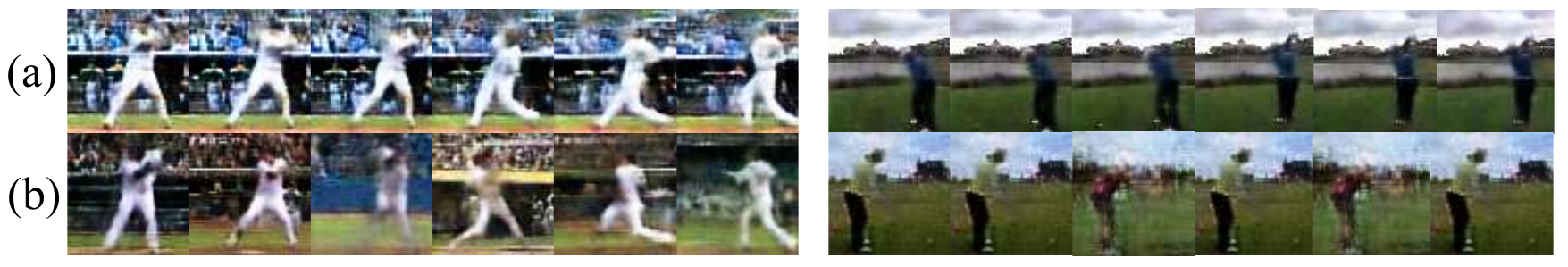}
\end{center}
	\vspace{-3mm}
   \caption{experimental results comparing with the NN method.
   (a) proposed method. 
   (b) NN with the latent space from VAE.
   }
\label{fig:exp_nn}
\vspace{-3mm}
\end{figure}
\noindent
\textbf{Parameter Fine-Tuning: }
After training the parameters $W_E, W_x$ and $W_D$ using the batch from the training dataset, we further fine-tune the parameters with the observed data pairs $(x_i,y_i),i=1,\cdots,N$ in the same way as previous regression techniques~\cite{rasmussen2006gaussian,lawrence2004gaussian}.
Note that the training of the regression part is not done because the ground truth is not available for the test dataset.
For the fine-tuning process, mini-batches are composed of the observed test data pairs $({x}_i,{y}_i)$ and randomly selected $(K-1)$ data sequences $(x^k_i,y^k_i)$ from the training set as in Fig.~\ref{fig:post_proc}, where $i=1,\cdots,N_k$ and $k=1,...,(K-1)$.
When the total number $N$ of observed test data pairs is less than $L$, we increase the number of samples by allowing repetition. 
Then, the parameters are fine-tuned with $50$ iterations. 
The detailed implementation is described in the supplementary material. 

%% file: 4_Experiment.tex
\section{Experiments}
In the experiments, we evaluated the regression capability of the proposed method via two applications composed of image data: 
(1) a problem with a simple temporal domain and complicated codomain and 
(2) a problem with a complicated domain and codomain.
For the first application, we used sport data sequences obtained from YouTube.
Human pose reconstruction for a given skeleton was tested for the second application.
\subsection{Sports Data Sequences}
\label{sec:sports}
\noindent
\textbf{Evaluation Scenario:} 
In this scenario, we created data sets for three sport sequences: baseball swing, golf swing, and weightlifting.
The dataset includes $236$ baseball swings, $232$ golf swings and $129$ weightlifting sequences from YouTube.
In the dataset, $1000-2000$ images are included for each action sequence, and their relative orders are given.
The domain is defined as $\mathcal{X}:[0, 1]$ and a point in $\mathcal{X}$ is assigned to $x$ for each image $y\in\mathcal{Y}$ according to its relative order in the entire sequence.
For testing, the golf and the baseball swings were trained with $200$ randomly selected sequences and tested with those that remained. 
The weight lifting scenario was trained with $100$ sequences.
We executed the regression for each test sequence with $20$ observed images within all images of each sequence and compared the results with multiple-output GP regression (MOGP)~\cite{alvarez2011computationally}, and GP regression combined with vanilla VAE~\cite{kingma2013auto} (called R-VAE from here on).
For R-VAE, we conducted the fine-tuning process in the same way as the proposed method. 
For MOGP, we trained the kernel with two-thirds of the images in the given sequences.

\noindent
\textbf{Qualitative Analysis:}
Fig.~\ref{fig:exp_sports_1} shows the qualitative comparison of image generation results. 
The sequences in Fig.~\ref{fig:exp_sports_1} show samples uniformly picked among the regressed responses from $100$ evenly divided points in the range $[0, 1]$. 
As seen in (a), the proposed method generated the most accurate responses compared to the other methods.
R-VAE also succeeded in capturing the blunt characteristics of the background and the motions of the actions.
However, the generated images in (b) suffer from large amount of noise for some images it is difficult to recognize the motion (circled in red).
Demonstrated are also instances in which the order of the image was not matched (circled in blue), and instances in which the background of the image was not matched (circled in green).
The images in the box show the samples of reconstruction results for given image pairs.
Both, the proposed method and R-VAE successfully reconstructed the images, but the regression performance is largely different.
As in (c), MOGP was not successful in describing the motion changes in the image, where every regression converged to the average of the training images.

\begin{figure}
\begin{center}
   \includegraphics[width=0.99\linewidth]{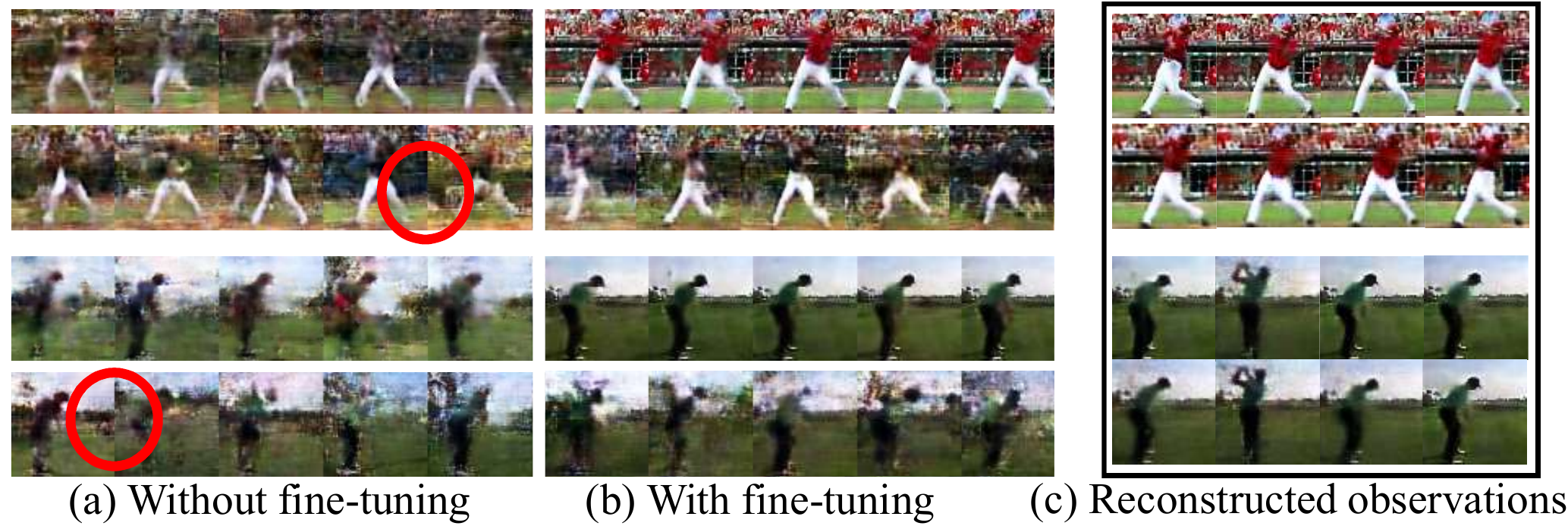}
\end{center}
	\vspace{-3mm}
   \caption{Analysis on the effect of fine-tuning.
   (a), (b): the regression result of the proposed method is shown in the first row and that of R-VAE in the second row.
   (c): the images in the box denotes the samples of reconstruction results for observed images.}
\label{fig:exp_sports_2}
\vspace{-3mm}
\end{figure}

We also conducted the experiments comparing with Nearest Neighbor (NN) method results to the proposed method.
We investigated the reconstruction results by applying the NN to the latent space after VAE learning.
However, the latent space encoded by the vanilla VAE is not appropriate enough to perform regression using the NN (see Fig.~\ref{fig:exp_nn}).
This is because the encoding of the background region plays a dominant role in NN as compared to the motion region. 
This problem is clearly seen at the bottom right sequence in the Fig.~\ref{fig:exp_nn}. 
Although the background (green and sky) region is relatively similar to the observation, the swinging human regions are not correctly regressed. 
This clearly shows that the proposed regression in the latent space is well performed achieving expected regression results in the image space. 
The encoder and decoder are trained to link the regression results in the latent space to the regression results in the image space directly, which is not trivial as shown in Fig.~\ref{fig:exp_nn} (b).
This is the second contribution of the proposed method (Abstract-ii). 

Fig.~\ref{fig:exp_sports_2} shows the effect of the fine-tuning process. 
The first and second column show the results with and without fine tuning.
The result of the proposed method is shown in the first row and that of R-VAE in the second row.
Before fine tuning, both methods generated noisy outputs, but the proposed method captured the vast characteristics of the background as well as the change of the motions.
In R-VAE, background information was less accurate than the proposed method (circled in red).
After the fine-tuning process, both methods accurately reconstructed the given image pairs, as in (c). 
Nevertheless, the regression performance between the methods varied significantly, as in (b).
\begin{figure}
\begin{center}
   \includegraphics[width=0.95\linewidth]{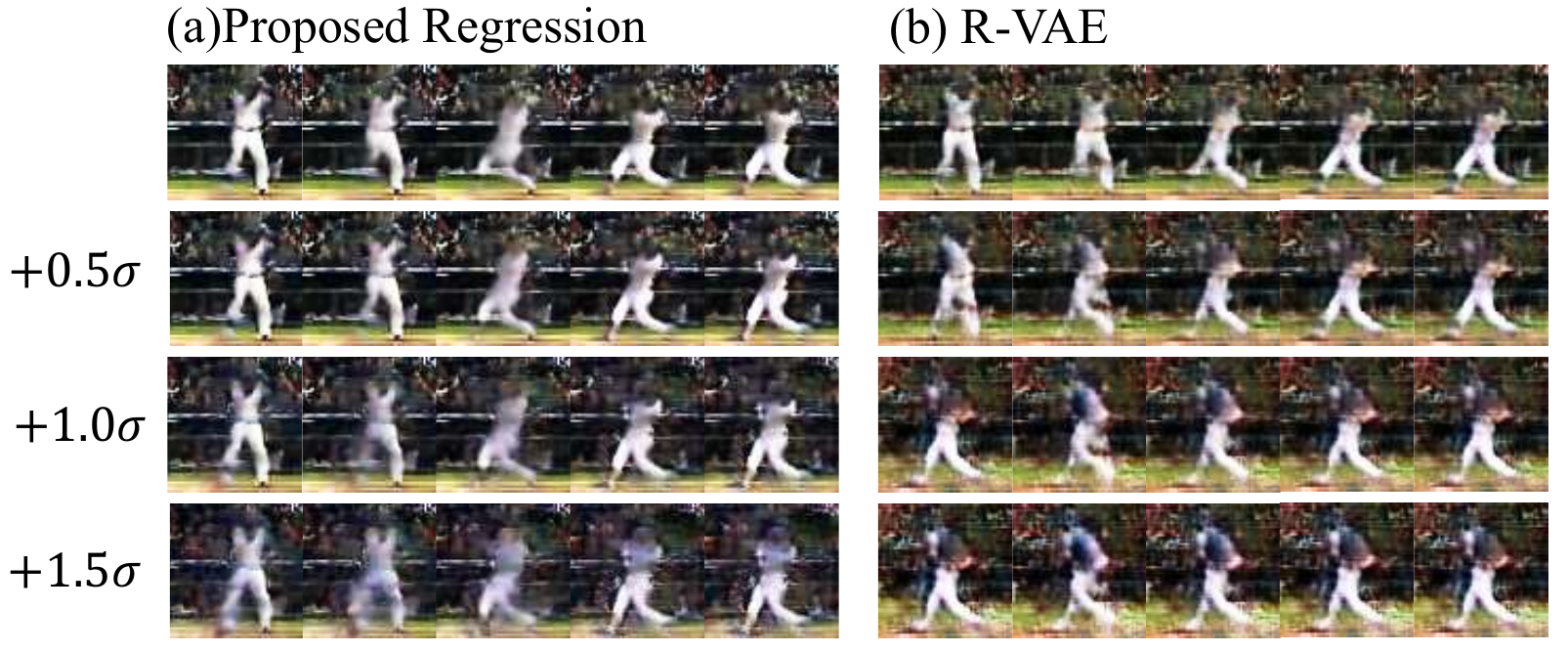}
\end{center}
\vspace{-2mm}
   \caption{Results from +$0.5\sigma, 1.0\sigma$ and $1.5\sigma$ latent sample.}
\label{fig:exp_sports_3}
\end{figure}

\begin{table}[t]
\centering
 \caption{Measure for the results with / without background.}
 \label{table:sport1}%
 \vspace{1mm}
 \resizebox{0.9\linewidth}{!}{
\begin{tabular}{|l|l|l|l|l|}
\hline  
\multicolumn{5}{|c|}{Structural Similarity Index Measure~\cite{wang2004image} result} \\  
\hline
sports& \textbf{Proposed}& \textbf{R-VAE~\cite{kingma2013auto}}& MIGP~\cite{alvarez2011computationally}& NN\\
\hline
Baseball&\textbf{0.610 / 0.607} & \textbf{0.492 / 0.489} & 0.803 / 0.247 & 0.215 / 0.210 \\
\hline
Golf &\textbf{0.752 / 0.707} & \textbf{0.578 / 0.543} & 0.845 / 0.114 & 0.244 / 0.213\\
\hline
Snatch &\textbf{0.377 / 0.369} & \textbf{0.207 / 0.205} & 0.626 / 0.019 & 0.206 / 0.198\\
\hline 
\end{tabular} 
}
\end{table}

Fig.~\ref{fig:exp_sports_3} represents the image generation results for different standard deviations.
As with the original GP regression, the proposed method estimates the output responses in the form of mean and variance because the latent $z$ for reconstructing the image is sampled from Gaussian distribution, as in~(\ref{eq:meanvar}).  
As seen in (a), the proposed algorithm captured the core semantics of the motion in each image despite the deviation change.
In R-VAE, the regression results were plausible when the sampled latent $z$ was close to the mean, but the motion in the image was regressed by a totally different action when adding large amounts of noise (up to 1.0$\sigma$).
From this result, we can see that R-VAE also has an ability to align the images in the latent space according to their order as reported in previous works~\cite{kingma2013auto,goroshin2015learning}.     
However, the results also show that the learned variance of R-VAE does not represent the motion semantics required for regression well, which is essential for the realization of GP regression in the image space.



\begin{table}[t]
\centering
 \caption{Measure for images from +$0.5\sigma, +1.0\sigma$ and $+1.5\sigma$.}
 \label{table:sport2}%
 \vspace{1mm}
  \resizebox{0.9\linewidth}{!}{
\begin{tabular}{|l|l|l|l|l|}
\hline  
\multicolumn{5}{|c|}{SSIM result for different standard deviations} \\ 
\hline
sports& method & +0.5$\sigma$& +1.0$\sigma$& +1.5$\sigma$\\
\hline
Baseball&proposed&0.6453 & 0.5980 & 0.5307 \\
&R-VAE~\cite{kingma2013auto}&0.4993 & 0.4402 & 0.3825 \\
\hline 
Golf&proposed&0.7203 & 0.4839 & 0.4422 \\
&R-VAE &0.5642 & 0.4026 & 0.2417 \\
\hline
Snatch&proposed&0.4042 & 0.3656 & 0.3629 \\
&R-VAE &0.2700 & 0.1645 & 0.0770 \\
\hline
\end{tabular}} 
\end{table}

\begin{figure*}
\begin{center}
   \includegraphics[width=0.95\linewidth]{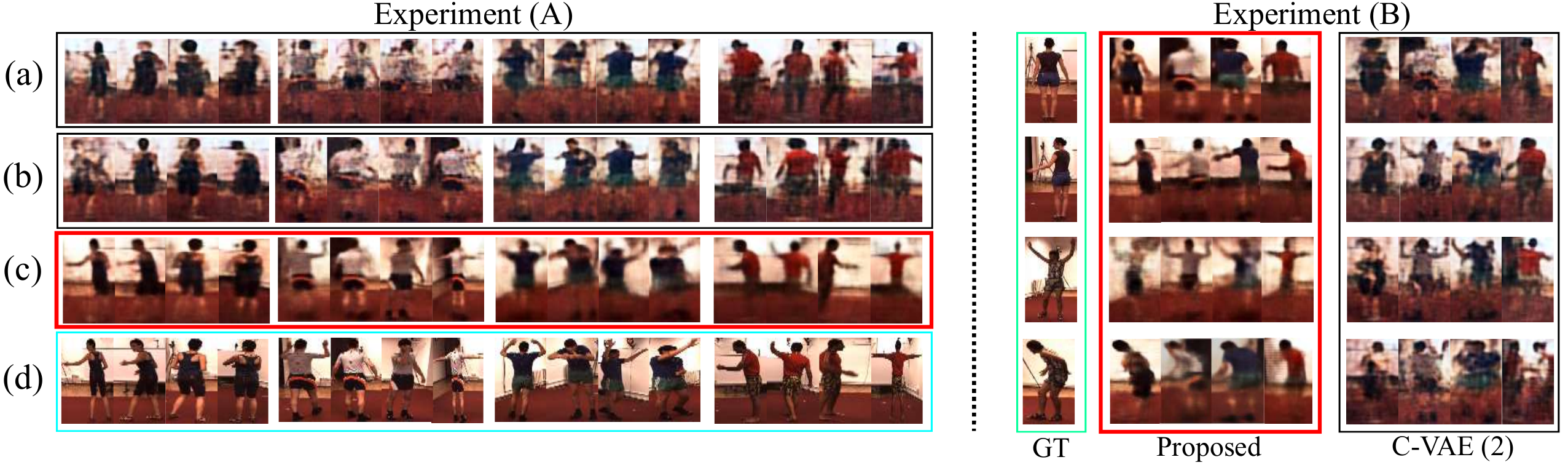}
\end{center}
\vspace{-2mm}
   \caption{Human pose estimation results from the joint (best viewed in color).
   The images in row (a) represents the C-VAE (1) results.
   The images in rows (b) is result from the CVAE (2). 
   The row (c) is the result from proposed method.
   The row (d) shows the ground truth.}
\label{fig:exp_h36m_1}
  \vspace{-2mm}
\end{figure*}

\noindent
\textbf{Quantitative Analysis:}
The quantitative performance was measured using the Structural Similarity Index Measure (SSIM)~\cite{wang2004image} which captures the structural similarities between two images. 
We estimated the $100$ images in the test set by using their domain information only, and compared the similarity between the ground truth image and the regression results. 
Table~\ref{table:sport1} shows the performance measures for generated regression images.
For the three different sport sequences, the proposed method generated more similar images to the ground truth (GT) compared to R-VAE.
Interestingly, the results of MOGP~\cite{alvarez2011computationally} which converged to the average of the images were measured to be most similar among the tested methods when including the background.
This is because the background of the average image is almost the same as the background of the GT when the background of the GT is fixed.
When we measured the similarity without the background region, MOGP was not successful and the proposed algorithm achieved the highest performance. 
Also, as with the result shown in Fig.~\ref{fig:exp_nn}, NN method marked unsuccessful quantitative performance.
Table~\ref{table:sport2} and Fig.~\ref{fig:exp_sports_3} show the performance when changing the standard deviation. 
We confirmed that the proposed method generated more plausible output than R-VAE for all cases.

\subsection{Human Pose Reconstruction}
\label{sec:h36m}
\noindent
\textbf{Evaluation Scenario:} 
For this experiment, we have used the human 3.6 million (H3.6m) \cite{h36m_pami} dataset for generating proper human appearance given the joint positions.
The dataset provides $32$ joint positions, and thus the input data lie in $96$ dimensional space.
The dataset includes diverse actions, and each action is repeatedly performed by different actors.
Our goal is to estimate the proper image of a new skeleton by utilizing the observed pairs of joint positions and images.
In the experiment, we used the `greeting' and `posing' scenarios of the H3.6m dataset. 
The scenario for each actor was captured in $8$ different view-points, resulting in a total of $16$ human pose sequences available for each actor.
We trained the model with the motions of $4$ different actors using $12$ sequences from each actor.
Then, we picked the observations from the remaining four sequences and conducted the regression.
The joint vectors for the regression were selected from the sequences from which the observations were selected. 
The joint vectors from other actors were also tested.
For comparison, we used the recent conditional VAE (C-VAE)~\cite{yan2015attribute2image} method, which generated an image according to a given attribute coupled with the sampled latent code.
In this experiment, the joint vector was used as the attribute.
\begin{table}[t]
\centering
 \caption{Similarity measure for generated human pose image.}
  \vspace{1mm}
 \label{table:human1}%
  \resizebox{0.9\linewidth}{!}{
\begin{tabular}{|p{1.2cm}|p{1.6cm}|p{1.3cm}|p{1.6cm}|p{1.2cm}|}
\hline  
\multicolumn{5}{|c|}{SSIM result for human pose generation} \\ 
\hline
Actors& proposed(A) & CVAE(A) & proposed(B) & CVAE(B)\\
\hline
\#1&0.7402&0.4849 & 0.5227 & 0.4059 \\
\#2&0.6743&0.4265 & 0.4775 & 0.3580 \\
\#3&0.7295&0.5094 & 0.5013 & 0.4268 \\
\#4&0.7671&0.4954 & 0.5224 & 0.4198 \\
\hline 
\end{tabular} }
\vspace{-2mm}
\end{table} 

\noindent
\textbf{Qualitative Analysis:}
Fig.~\ref{fig:exp_h36m_1} (A) shows the pose generation result of the proposed algorithm and C-VAE.  
For C-VAE $(1)$, we used randomly sample latent code $z_y$ as in~\cite{yan2015attribute2image}.
For C-VAE $(2)$, the latent code was given by the proposed regression block in Fig.~\ref{fig:overview}. 
As shown in (c), the image regressed by the proposed method successfully describes the overall motion of each human pose.
Also, note that the background of each image was correctly generated according to the view point of the observed data pair. 
The generated images from C-VAE $(2)$ contain a large amount of noise, but they captured the rough silhouette of the actors.
This result is noticeable because C-VAE usually deals with cases in which the attribute is discrete.
The result from C-VAE $(2)$ was clearer than the result from C-VAE $(1)$, but the difference was not significant.
The result in Fig.~\ref{fig:exp_h36m_1} (B) shows the output responses when the joint vectors of other actors were given.
The images in the blue box refer to the ground truth pose, and the images in the red box are the regression result by the proposed method.
This result shows that the proposed method generates poses that resemble those of the input joint vectors while preserving the appearance of the given data pairs via regression.  
Specifically, when the given pair involves a man wearing white clothes, the generated image illustrates a man wearing the same clothes with a similar pose to the GT image.
C-VAE (2) was not successful in generating a corresponding pose for a given joint from other actors.

\noindent
\textbf{Quantitative Analysis:}
Table~\ref{table:human1} shows the similarities between the generated image and the ground truth image.
The first two columns denoted by (A) represent the quantitative results in experiment (A) of Fig.~\ref{fig:exp_h36m_1}.
In the experiment, the proposed method achieved a higher score than C-VAE (2).
For experiment (B), we compared the similarity between the regressed image and the original images for the joint vector (green box in Fig.~\ref{fig:exp_h36m_1}).
There, our method also achieved a higher score than C-VAE (2).

In this experiment, the input data lay in the high dimensional space and the target joint vectors were selected without considering temporal information.
Despite the complicated and non-sequential input domain, the proposed regression method achieved reasonable output responses describing the semantics given in the input and  the identity information contained in the observed pairs.
It means that the proposed method is available for the temporal input and can also handle more complex and non-sequential input.

%% file: 5_Conclusion.tex
\section{Conclusion}
In this paper, we have proposed a novel regression method regarding high dimensional visual output.
To tackle the challenge, the proposed regression method is designed so that the result of the regressed response in a latent space and the corresponding response in the data space coincide.
Through qualitative and quantitative analysis, it has been verified that our method properly estimates the regressed image responses and offers an approximation of the complicated input-output relationship.
This paper discovers meaningful progress in the regression field in that our work introduces a way to combine a deep layered architecture to the regression method in a probabilistic framework.

%% file: 6_Appendix.tex
\clearpage
\begin{appendices}

\begin{figure}
\begin{center}
   \includegraphics[width=0.99\linewidth]{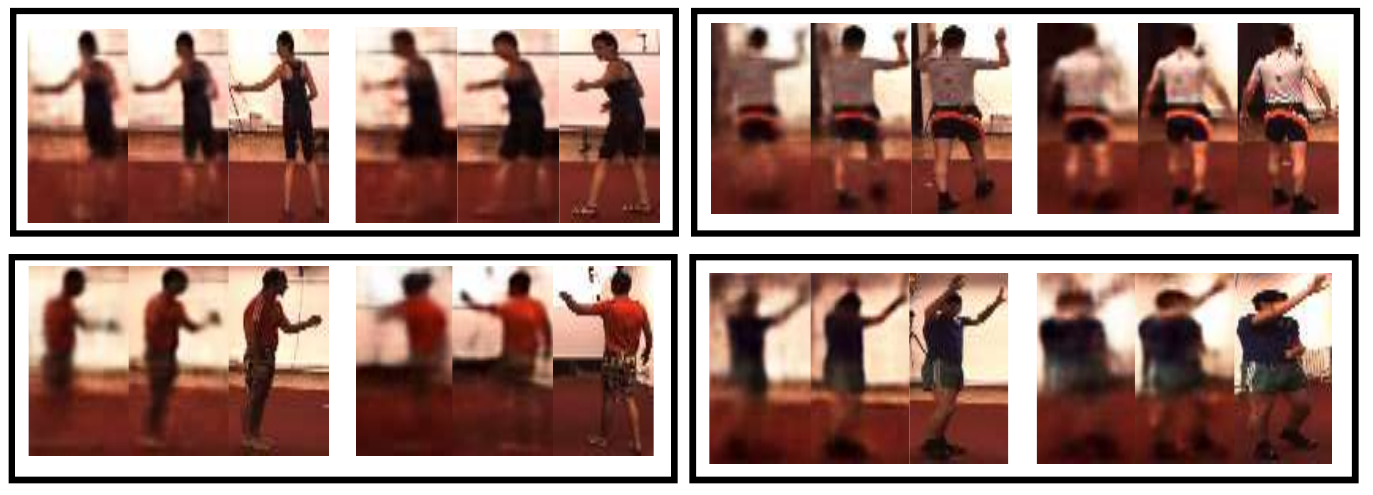}   
\end{center}
   \caption{Regression and reconstruction result from the same data pair. 
   Each image set is composed of three images. Within each set, the leftmost image is generated from a regression; 
   the middle image refers to the result from reconstruction using the joint vector and the corresponding image; and
   the rightmost image shows the ground truth.}
\label{fig:supple_human}
\end{figure}

\begin{figure}
\begin{center}
   \includegraphics[width=0.99\linewidth]{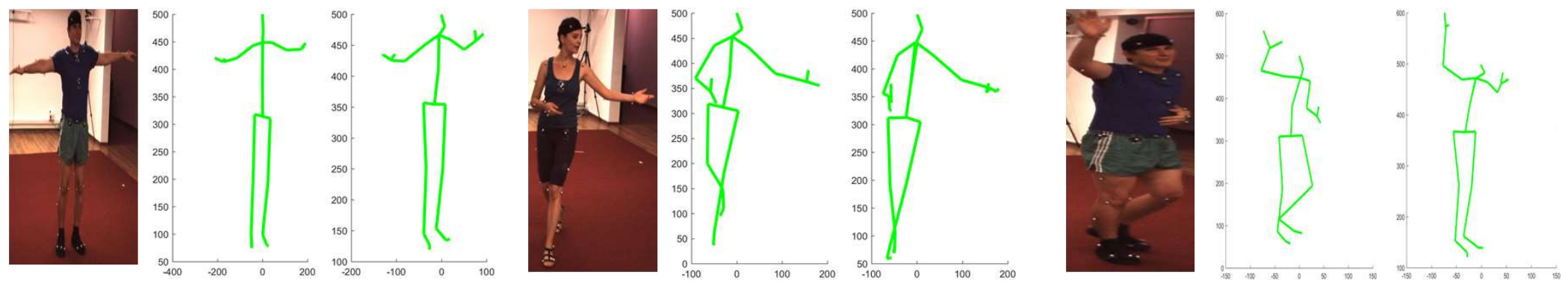}   
\end{center}
   \caption{ left: input image, center: ground truth, right: regressed result.}
\label{fig:exp_joint}
\end{figure}

\section{Implementation Detail}

In the experiments, the encoder $[m_{i,y},\sigma_{i,y}] = E(y_i;W_E)$, decoder $D(z;W_D)$ and $\sigma(y;W_E)$ of the kernel for GP regression in Figure~2 of the paper are defined as multi-layered perceptrons.
The encoder $E(y;W_E)$ is designed with five convolution layers and one fully connected layer (the convolution layers are composed of $16, 32, 64, 128, 256$ channels with filter size $5\times 5$ each).
All $y\in\mathcal{Y}$ are resized to three channel $64$-by-$64$ images. We set the dimension of the $m_{i,y}$ and $\sigma_{i,y}$ to $128$, and the fully connected layer returns the $256$ elements for $m_{i,y}$ and $\sigma_{i,y}$.
The former $128$ elements are used as $m_{i,y}$, and the latter $128$ entries are defined as $\sigma_{i,y}$.
For the mapping function $[m_{i,x}, \sigma_{i,x}]=f(x_i,W_x)$, $m_{i,x}$ refers to $x_i\in\mathcal{R}^{n(\mathcal{X})}$ and the additional $n(\mathcal{X})$ outputs in $E(y;W_E)$ indicates $\sigma_{i,x}$, as in Figure 2 of the paper.
Therefore, the overall dimension of the final fully connected layer is $256+n(\mathcal{X})+1$; $256$ dimensions for $[m_{i,y}, \sigma_{i,y}]$, $n(\mathcal{X})$ dimensions for $\sigma_{i,x}$, and one dimension for $\sigma_k$.
For the decoding function $\hat{y}=D(z;W_D)$, $6$ convolution layers with $2$-by-$2$ upsampling are used to reconstruct the image. 
The convolution layers have 256, 128, 64, 32, 16, 3 channels with filter size $4\times4, 5\times5, 5\times5, 5\times5, 5\times5$ and $5\times5$.
\section{Additional Results}

\begin{figure}
\begin{center}
   \includegraphics[width=0.99\linewidth]{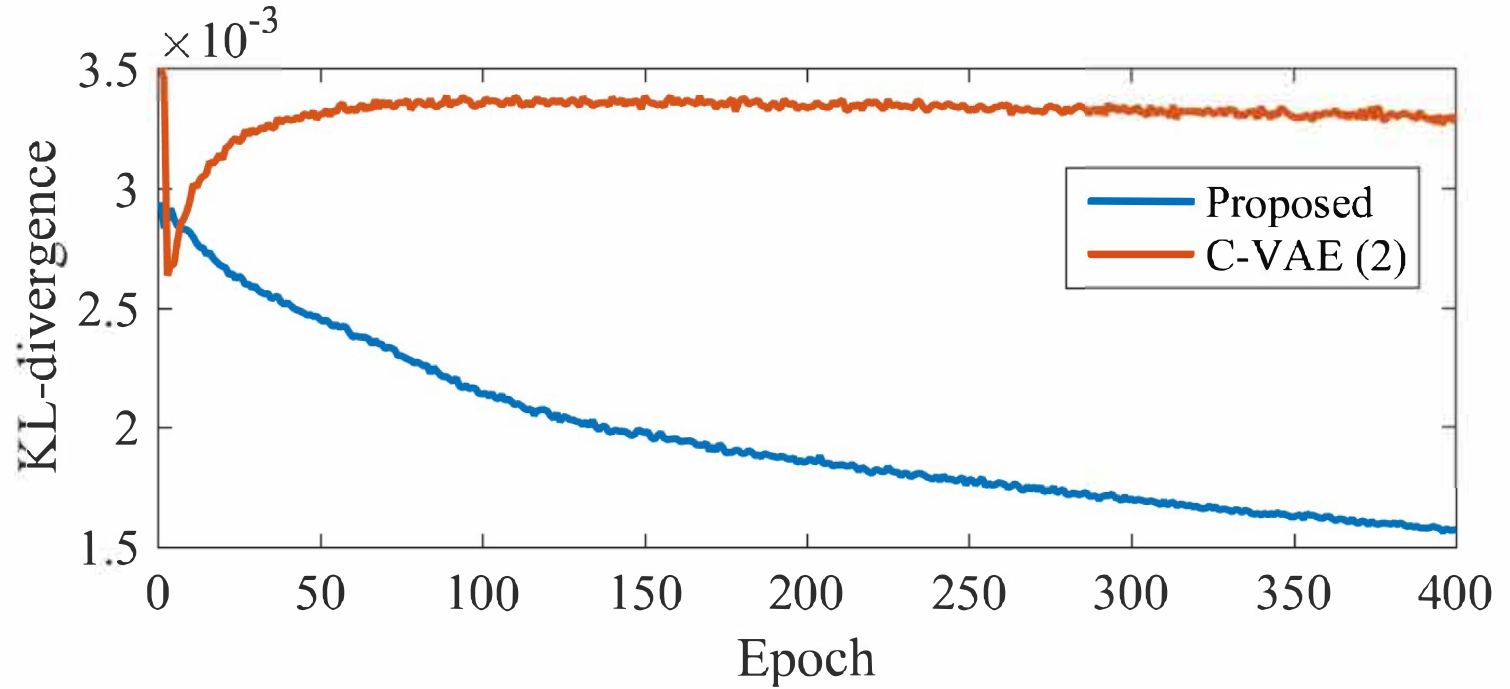}   
\end{center}
   \caption{KL divergence between the latent distributions for regression and reconstruction, from the same joint vectors.}
\label{fig:supple_human_1}
\end{figure}

To check the validity of the regression procedure conducted in a latent space, 
we compared the latent vector obtained by regression with the latent vector obtained by reconstruction for an input data pair.
For the reconstruction, we used both the joint vector and the corresponding image in the H3.6m dataset~\cite{h36m_pami}.
For the regression, only the joint vector was given and the projected point was estimated by the proposed regression method.
Since the latent vectors were obtained from the same input data,  in ideal conditions the vectors should converge to the same location.
Figure~\ref{fig:supple_human} shows the qualitative results for the regression and the reconstruction for the same input data pair, where it can be seen that
both responses converged to the ground truth image.
The graph in Figure~\ref{fig:supple_human_1} indicates the KL-divergence between the two latent vectors.

\begin{figure}[t]
\begin{center}
   \includegraphics[width=0.99\linewidth]{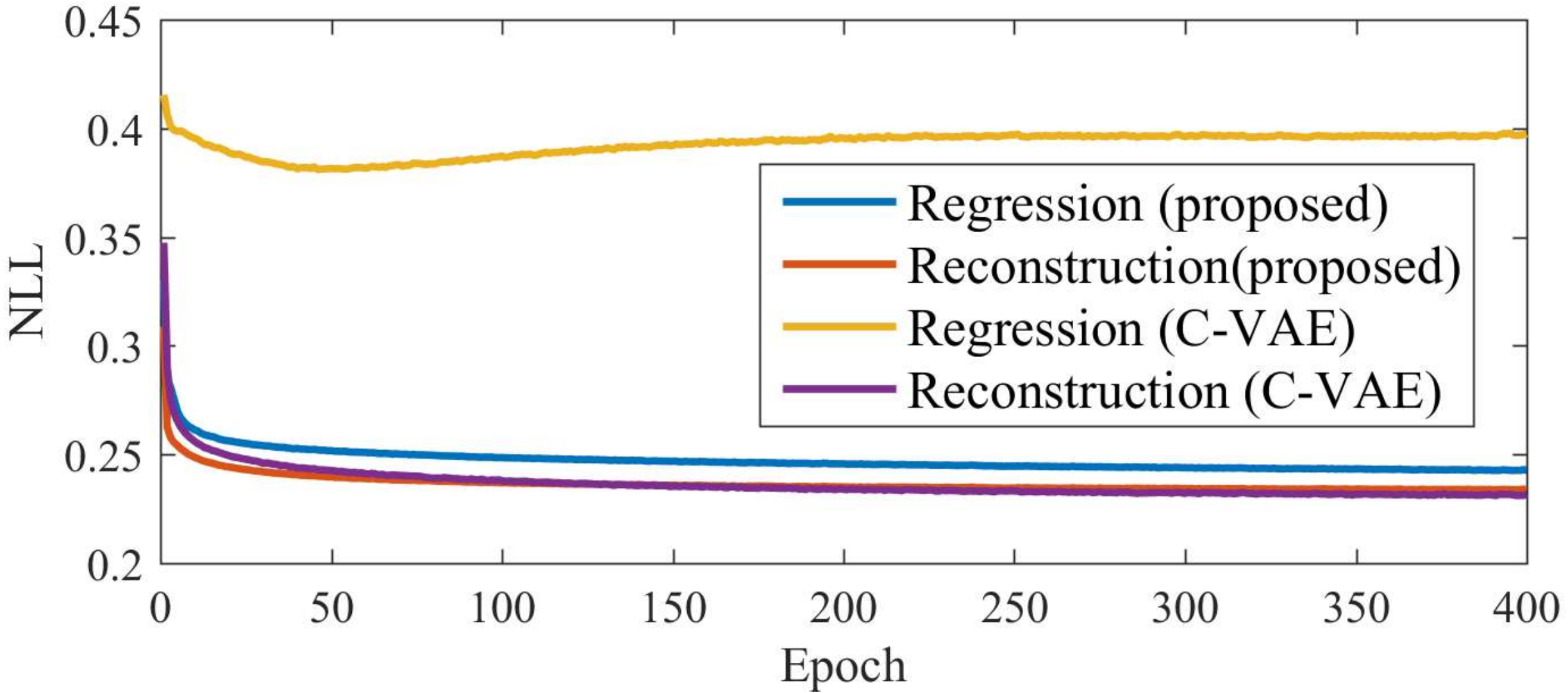}   
\end{center}
   \caption{Negative log likelihood ratio for the regressed and reconstructed visual responses, from the proposed method and C-VAE (2).}
\label{fig:supple_human_2}
\end{figure}

Since the latent vector $z$ in the paper is defined by a Gaussian distribution, we used the KL-divergence as distance measure.
As seen in the graph, the KL-divergence obtained by the proposed method was gradually decreased.
The result demonstrates that the two vectors obtained from both cases converged to the same location, as expected.
When tested with the C-VAE (2), the divergence did not converge.

As shown in Figure~\ref{fig:supple_human_2}, we also measured the changes of the negative log likelihood (NLL), for both the regression and the reconstruction.
In the proposed method, we confirmed that the NLL ratio for both cases converged.  
In C-VAE (2), the NLL ratio converged only when both the joint vector and the images were given, but it did not successfully converge when the regression was applied.

\begin{figure*}
\begin{center}
   \includegraphics[width=0.99\linewidth]{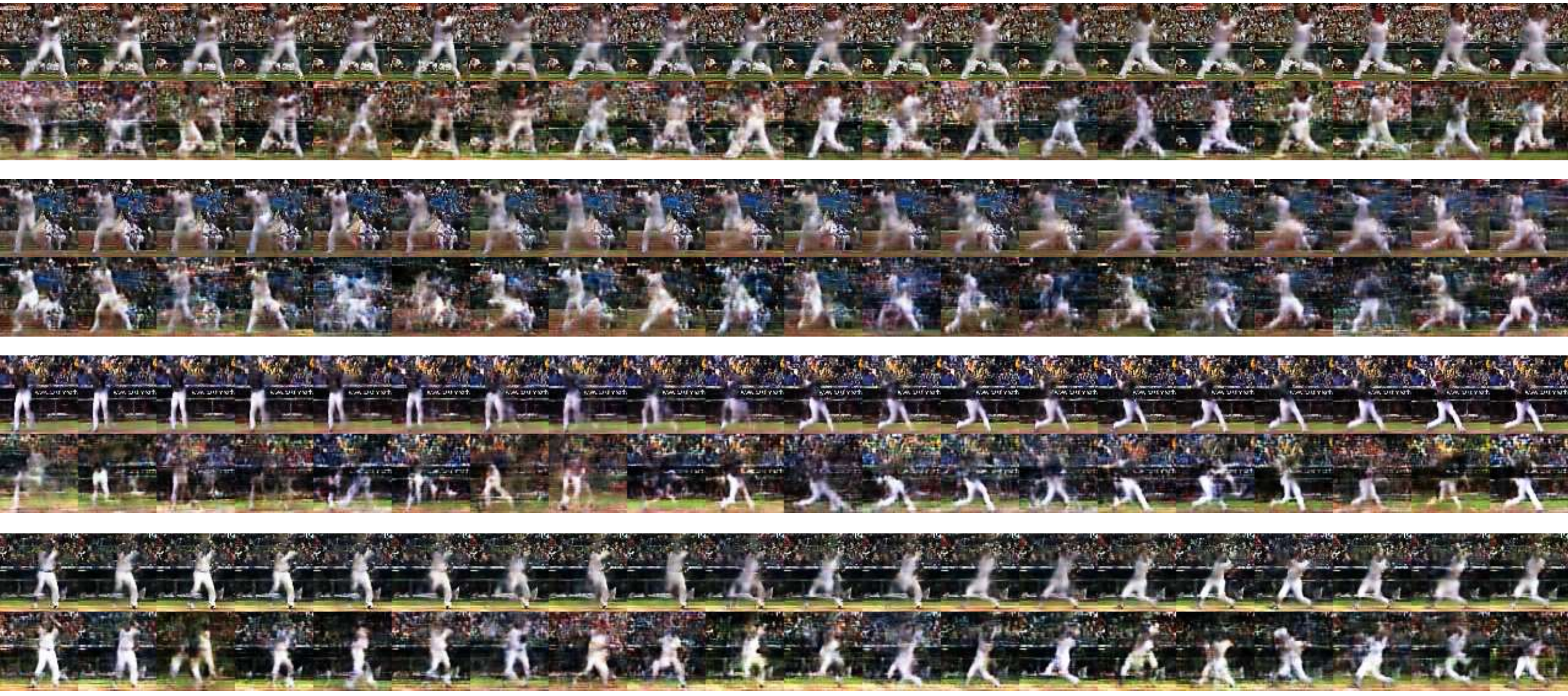}   
\end{center}
   \caption{Qualitative results on regression from the baseball swing dataset. The first row in each action represents the proposed method, and the second row shows the result from R-VAE.}
\label{fig:supple_baseball}
\end{figure*}

\begin{figure*}
\begin{center}
   \includegraphics[width=0.99\linewidth]{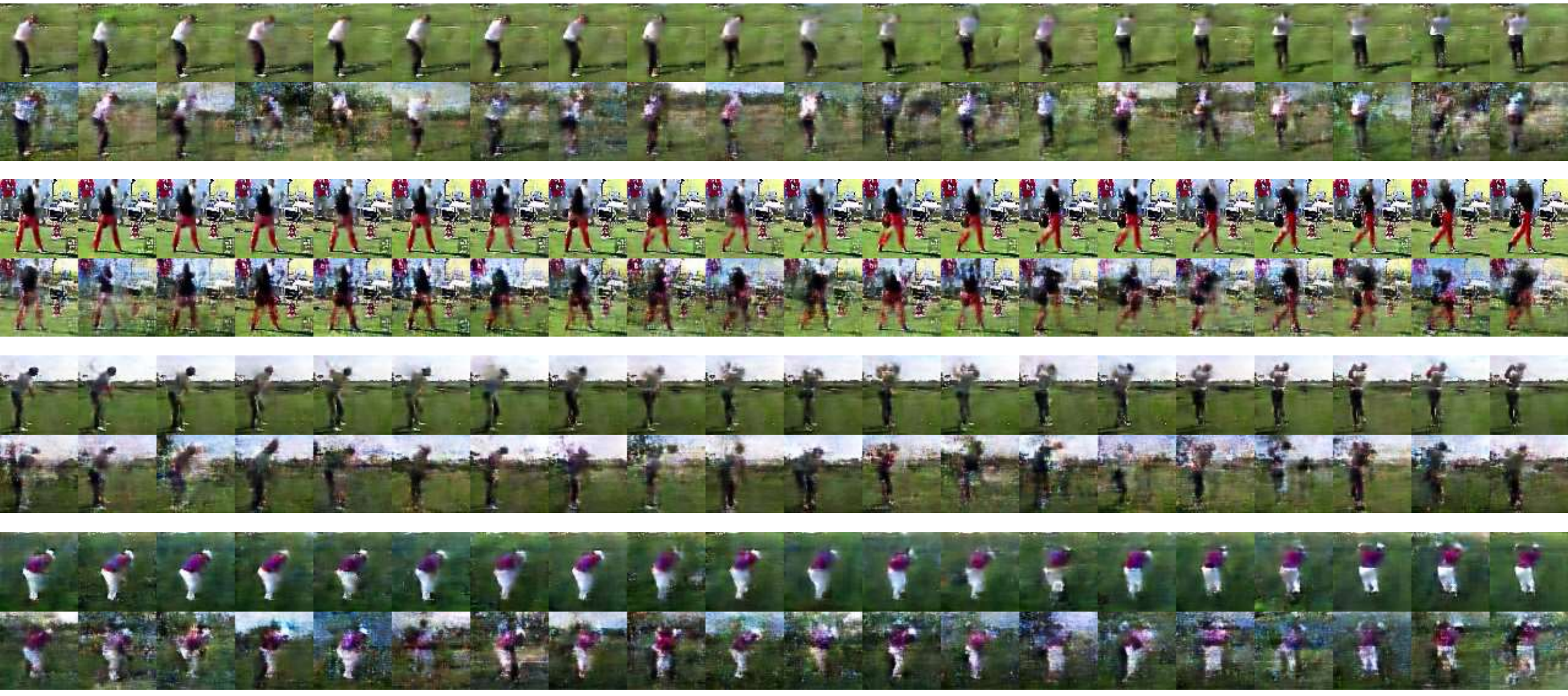}   
\end{center}
   \caption{Qualitative results on regression from the golf swing dataset. The first row in each action represents the proposed method, and the second row shows the result from R-VAE.}
\label{fig:supple_golf}
\end{figure*}

\begin{figure*}
\begin{center}
   \includegraphics[width=0.99\linewidth]{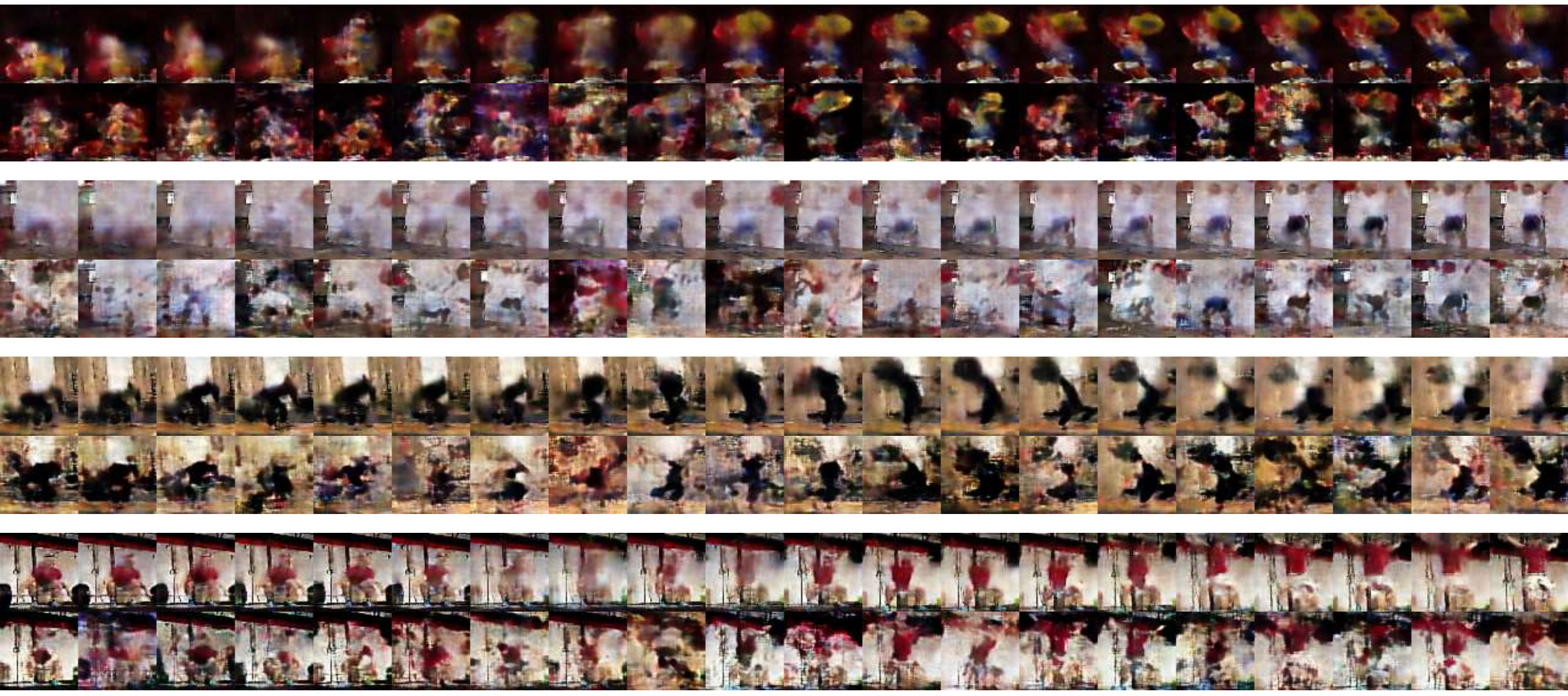}   
\end{center}
   \caption{Qualitative results on regression from the weightlifting dataset. The first row in each action represents the proposed method, and the second row shows the result from R-VAE.}
\label{fig:supple_snatch}
\end{figure*}

Figure~\ref{fig:supple_baseball}, Figure~\ref{fig:supple_golf} and Figure~\ref{fig:supple_snatch} show additional generation results of sports sequences.  
The figures describe the regression results of the proposed method and of R-VAE in our work, which are the supplementary results of Figure 7 included in the submitted version.
We confirmed that the proposed method achieved a superior regression performance for diverse action sequences compared to R-VAE.

In addition to the image sequence regression and human pose reconstruction examples, we newly performed human joint estimation experiments as shown in Fig.~\ref{fig:exp_joint}.
Although our method is not originally designed for the human joint estimation purpose, the proposed method could successfully estimate the human joints by performing regression using an (image - joint) data pair and finding the corresponding joints when a new image is given.
This example further shows that the proposed method is applicable to practical applications composed of various data pairs.


\end{appendices}